\documentclass[sigconf, anonymous=false]{acmart}
\AtBeginDocument{%
  \providecommand\BibTeX{{%
    \normalfont B\kern-0.5em{\scshape i\kern-0.25em b}\kern-0.8em\TeX}}}
    
\usepackage{algorithm}
\usepackage{algorithmic}

\setcopyright{none} 
\renewcommand\footnotetextcopyrightpermission[1]{}
\begin{document}

\title{Exploration-Exploitation Motivated Variational Auto-encoder for Recommender Systems}

\author{Yizi Zhang}
\affiliation{
  \institution{Duke University}
  \city{Durham}
  \state{NC}
  \country{USA}
 }
\email{yizi.zhang@duke.edu}

\author{Meimei Liu}
\affiliation{
  \institution{Virginia Tech}
  \city{Blacksburg}
  \state{VA}
  \country{USA}
 }
\email{meimeiliu@vt.edu}
\begin{abstract}
  Recent years have witnessed rapid developments on collaborative filtering techniques for improving the performance of recommender systems due to the growing need of companies to help users discover new and relevant items. However, the majority of existing literature focuses on delivering items which match the user model learned from users' past preferences. A good recommendation model is expected to recommend items that are known to enjoy and items that are novel to try. In this work, we introduce an exploitation-exploration motivated variational auto-encoder (\textbf{XploVAE}) to collaborative filtering. To facilitate personalized recommendations, we construct user-specific subgraphs, which contain the first-order proximity capturing observed user-item interactions for exploitation and the high-order proximity for exploration. A hierarchical latent space model is utilized to learn the personalized item embedding for a given user, along with the population distribution of all user subgraphs. Finally, experimental results on various real-world datasets clearly demonstrate the effectiveness of our proposed model on leveraging the exploitation and exploration recommendation tasks.
\end{abstract}




\maketitle
\pagestyle{plain}

\section{Introduction}
Recommender systems have been widely adopted by many online services, including commercial platforms and social media sites. For example, in e-commerce recommender systems, one of the overarching goals is to find the best products for each customer to fit their specific interests and needs. Given the explosive growth of information available on the web, personalized recommendation is an essential demand for facilitating a better user experience. Another important need of recommender systems is the exploitation-exploration, \textit{i.e.,} being able to exploit the observed user-item behaviors and explore the unobserved user-item interactions. 

There have been substantial methods proposed on link prediction in the social recommendations literature.  One of the most popular approaches is collaborative filtering (CF) \cite{sarwar2001item,schafer2007collaborative} that aims to model the users' preferences on items based on their previous interactions (\textit{e.g.,} ratings, clicks, purchases). Reinforcement learning is another efficient direction in dealing with recommender systems especially focusing on the trade-off between exploitation and exploration \cite{yogeswaran2012reinforcement,bouneffouf2012contextual}. 
Variational auto-encoder (VAE) \cite{kingma2014adam} has been applied to recommender systems recently. VAE  
is essentially a generative model incorporated within
a deep neural network. 
It models the population distribution of the input data through a simple distribution for the latent variables combined with a complex nonlinear mapping function. 
Recently,
lots of efforts have been devoted to improving the capability of
preserving the high-order proximity within the networks in an explicit or implicit way, such as \cite{grover2016node2vec,perozzi2014deepwalk,tang2015line,ou2016hope,qiu2017netmf}.
However, existing VAE-based recommendation models do not consider the high-order proximity, since the target function is to recover the explicit relationships between users and items; see Section \ref{sec:related} for details. 

In this paper, we propose a novel VAE-based approach for personalized recommender system to address the aforementioned gap in leveraging both exploitation and exploration. We denote our method as XploVAE, representing exploitation-exploration motivated VAE. We propose to model the high-order connectivity information via the construction of user-specific subgraphs, comprising of the first-order proximity that captures the observed user-item links and the high-order proximity for implicit user-to-item relations. For the purpose of exploration, we introduce transitivity to the bipartite graph to expressively model the high-order proximity as the unobserved but transitive links between users and items. A hierarchical latent space model is then integrated into a VAE-framework to learn the personalized item embedding of a given user subgraph, along with the population distribution of all user subgraphs. By aggregating the embeddings of the interacted items, we also enforce the embeddings to leverage the collaborative signals in item-item similarity.

The main contributions of this work are summarized as follows:
\begin{itemize}
    \item We highlight the importance of explicitly modeling the high-order proximity in the embedding functions of VAE-based methods. The proposed user-specific subgraphs contain first-order connectivity capturing observed user-item interactions for exploitation of known user preferences and high-order connectivity for exploration of novel products.
    \item Compared to having item embeddings shared by all users, our method learns personalized item embeddings to recommend more relevant items to the target user. With the hierarchical latent space model, we also learn the population distribution of all users to infer the aggregate user preferences.
    \item We effectively encode collaborative signals in the item embeddings to leverage the item-to-item similarity. 
\end{itemize}
Empirically, our proposed method outperforms state-of-the-art baselines on three benchmark datasets, including a recently proposed VAE-based method and several methods that either preserve the high-order proximity or consider exploration of novel items.

\section{Related Work}\label{sec:related}

We review existing work on representation-based models, VAE-based methods, exploitation-exploration motivated approaches and graph convolutional networks, which are most relevant with this work. 

Representation-based models have sparked a surge of interest since
the Netflix Prize competition \cite{bennett2007netflix} demonstrates matrix factorization models are superior to classic
neighborhood methods for recommendations. After that, various methods \cite{koren2009matrix, koren2015advances} have been proposed to learn the representations of users and items for better estimating the users’ preferences on items. Latent factor models are the mainstream among various collaborative filtering techniques due to their simplicity and effectiveness. They measure the interactions between users and items by multiplicating their latent features linearly; see \cite{koren2008factorization,agarwal2009regression} and reference therein. However, \cite{he2017neural} has shown that such inner product may not be sufficient to capture the complex structure of user behaviors. In XploVAE, we couple the proposed hierarchical latent space model with deep neural networks to improve its efficiency in dealing with complex data. 

Some recent methods use the auto-encoder (AE) or VAE to learn the item-based or user-based embedding separately by using the preference matrix. Hybrid VAE in \cite{gupta2018hybrid} uses VAE to reproduce the whole users’ preference history. 
\cite{liang2018variational} proposes a neural generative model to characterize a user’s preferences with a multinomial likelihood conditional on a latent user representation learned through VAE. However, the above methods do not consider exploration. In XploVAE, we exploit the high-order connectivity for exploration purposes.

Most recommender algorithms produce types similar to those
the target user has accessed before. This is because they
measure user similarity only from the co-rating behaviors
against items and compute recommendations by analyzing
the items possessed by the users most similar to the given
user. Incorporating exploration into recommender systems is of critical importantance - without it new or unseen items don’t stand a chance against previously accepted or more familiar items. One exploration-exploitation approach is via the $\epsilon$-greedy algorithm, where the loss function allocates $\epsilon$ percents component to explore new items in a random manner; the remaining components are reserved for exploitation; see \cite{chen2018fast}. Another approach is based on the upper confidence bound (UCB) that constructs confidence bounds associated with each item to capture how uncertain we are about the items; see \cite{auer2002using,vanchinathan2014explore}. Recent studies regard the problem as multi-armed bandits or contextual bandits, and solve it with
intricate exploration policies \cite{li2010contextual, McInerney2018mab}. However, these approaches are usually computationally intractable for nonlinear models, which terrifically limits its usage in recent advanced deep models \cite{cheng2016wide}. Rather than designing the sophisticated exploration strategies, our approach considers exploration of unobserved user-item interactions via constructing the high-order proximity between users and items. To model the high-order proximity, we apply an associative retrieval framework \cite{huang2004associative} to explore transitive associations among users through their past interactions. Such transitive associations are a valuable source of information
to help infer user interests and can be exploited to deal with the data sparsity problem. 

In addition to modeling the high-order proximity, we employ graph convolution operations and incorporate structured external information (\textit{e.g.,} the social relationship among
users or features related to items ) to explicitly encode the crucial collaborative signals (\textit{i.e.,} geometric structure) to improve the user and item representations. Many existing recommendation methods\cite{xiang2019ngcf, yu2020graph} also apply the graph convolution network (GCN) \cite{kipf2016gcn} on user-item graph. For example, GC-MC \cite{gcmc} employs one convolutional
layer to exploit the direct connections between users and
items. However, it fails to reveal collaborative signals in high-order
connectivities with only one convolutional layer. PinSage\cite{ying2018graph} employs multiple graph convolution layers on item-item graph for image recommendation. As such, the collaborative signal is captured on the level of item relations. We devise a specialized graph convolution operation on the item embeddings in a similar fashion to leverage item-item similarity.

\section{Problem Scenario}\label{sec:pre}
 We first introduce some notations and definitions. 
Let $\boldsymbol{G}= (\boldsymbol{U}, \boldsymbol{V}, \boldsymbol{E})$ be a bipartite graph, where the set of vertices $\boldsymbol{U} = (u_1, \cdots, u_{|\boldsymbol{U}|})$ denotes users, and $\boldsymbol{V} = (v_1, \cdots, v_{|\boldsymbol{V}|})$ denotes the set of items. $\boldsymbol{E}$ is an $|\boldsymbol{U}|\times |\boldsymbol{V}|$ incidence matrix, with entries $E_{ij}$ depicting explicit interactions between $u_i$ and $v_j$.  
 For example, in e-commercial recommender systems, 
 $E_{ij} =0,1,2,3$ representing no access, click, add to cart and buy actions, respectively. 
In rating-based recommender systems, 
$E_{ij}=0,1,\cdots, 5$ means the possible ratings. 
In this work, we consider $E_{ij}=1$ or $0$ representing access or no access. We call $E_{ij}$ the \textit{first-order proximity} that captures the observed user-item links, and can be used for exploitation. 
For the purpose of exploration, we further introduce a \textit{high-order proximity} for implicit
relations, i.e., the unobserved but transitive links, between users and items. 

\begin{figure}
\centering
\includegraphics[width=0.3\textwidth]{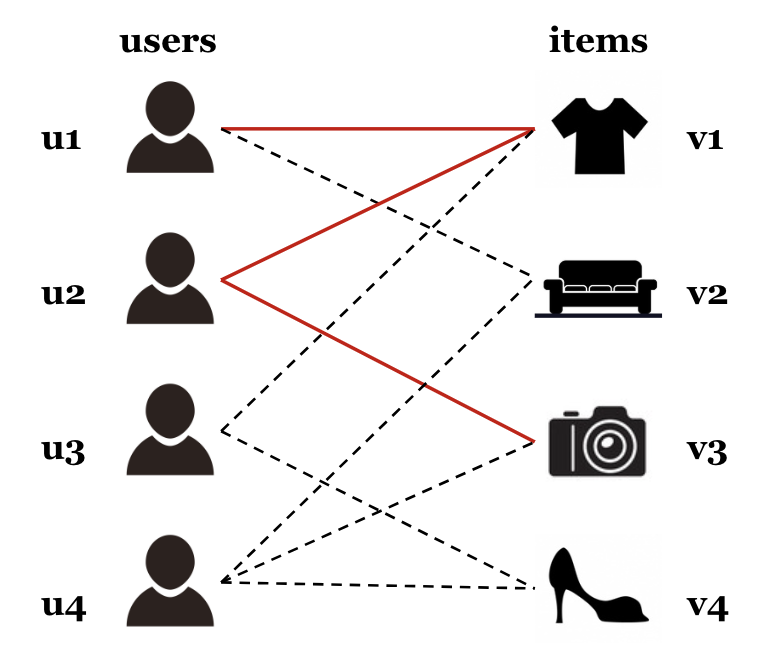}
\caption{Second-order proximity (red solid line) for implicit user-item interactions. The nodes denote users and items, while the edges (dashed lines) represent access or no access.}\label{fig:k_order_prox}
\end{figure}

\begin{definition}{\textit{\textbf{(k-th order proximity)}}} 
\textit{
For user $u$ and item $v$, denote the $k$-th order proximity between $u$ and $v$ as $E_{uv}^{(k)}$. The \textit{first-order proximity} $E_{uv}^{(1)}= E_{uv}$. For $k\geq 2$, $E_{uv}^{(k)}$ is the number of paths of length $k$ connecting $u$ and $v$. 
If no such a path, then $E_{uv}^{(k)}=0$.}
\end{definition}
By the definition, the $k$-th order proximity between user $u$ and item $v$ can be explicitly calculated as the $uv$-th entry $E_{uv}^{(k)}$ of the matrix
\begin{equation*}
    \boldsymbol{E}^{(k)} = \underbrace{\boldsymbol{E} \boldsymbol{E}' \cdots \boldsymbol{E} \boldsymbol{E}' }_{2(k-1)+1}.
\end{equation*}
Figure \ref{fig:k_order_prox} shows an illustrative example. There exists a $2$-step path between $u_1$ and  $v_3$ through $\textrm{user}\; 1 
{\to}\; \textrm{item}\; 1
{\to}\;\textrm{user}\; 2
{\to}\;\textrm{item}\; 3.$ The total number of such length-2 paths between user $u$ and item $v$ is the second-order proximity between the two vertices. 

For each user $u$, we construct a subgraph $\boldsymbol{A}^{(u)}\in \mathbb{R}^{K \times |\boldsymbol{V}|}$ based on the $k$-th order proximity ($k=1,\dots, K$) as follows.  

\begin{definition}{\textit{\textbf{(user subgraph)}}} \label{def: subgraph}
\textit{
A user subgraph $\boldsymbol{A}^{(u)} \in \mathbb{R}^{K \times |\boldsymbol{V}|}$, where $K$ is the total order of proximity. The $kv$-th entry $A^{(u)}_{kv}$ is defined as 
$$
A^{(u)}_{kv}: = 
    \begin{cases}
      1 & \text{if $E_{uv}^{(k)}\geq c_k$}\\
      0 & \text{otherwise}
    \end{cases},
$$
where $c_1=1$, and $c_k$ is a constant threshold to be specified.
}
\end{definition}
For user $u$ with corresponding subgraph $\boldsymbol{A}^{(u)}$, the edge  $A^{(u)}_{kv}$ represents whether user $u$ has access or not to the $v$-th item in the $k$-th order proximity. 
The first row of $\boldsymbol{A}^{(u)}$ characterizes the explicit links between the user $u$ and items in $V$, while the rows left represent implicit links via the high-order proximity. In practice, we set $K=2$, i.e., we only consider the first-order and the second-order proximities when constructing the user subgraphs, since exploring a higher-order proximity beyond the second-order is computationally expensive and renders inference on the user subgraph less reliable.

\section{Method}\label{sec:method}

We propose a novel approach by incorporating a hierarchical latent space model into the VAE framework. The goal is to learn latent embeddings of individual user subgraphs and the population distribution of all subgraphs for link prediction. XploVAE relies on VAE which consists of two components. The first component is a generative model that specifies how the latent variable $\boldsymbol{z}_u$ gives rise to the observation $\boldsymbol{A}^{(u)}$ through a nonlinear mapping parametrized by neural networks. The second
component is an inference model that learns the inverse mapping from $\boldsymbol{A}^{(u)}$ to $\boldsymbol{z}_u$. 

\begin{figure*}[ht!]
\centering
\includegraphics[width=0.78\textwidth]{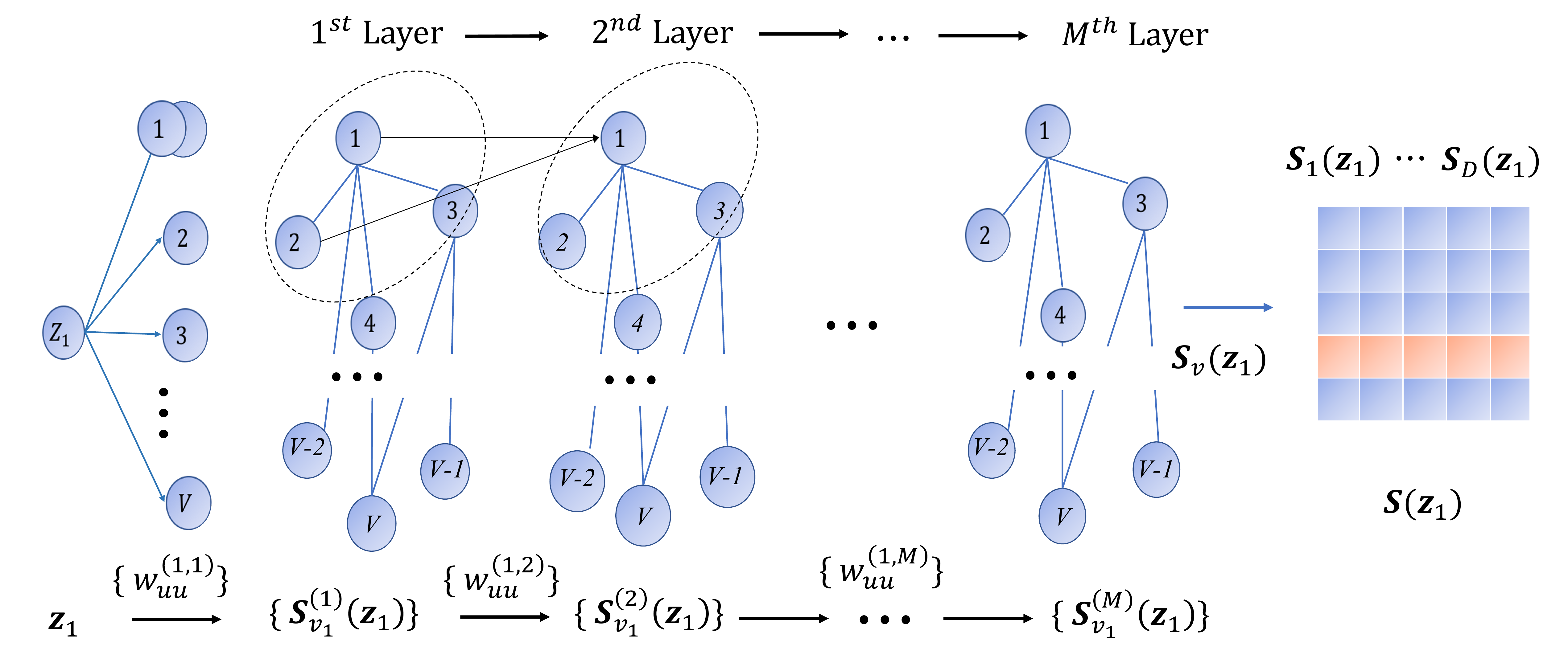}
\caption{Illustrative example of a $M$-layer neural network with 2-NN neighborhood for learning $\boldsymbol{s}_v(\boldsymbol{z}_u)$.  For example, for node 1,
the 2-NN is node 2 and node 3. For user $u = 1$ and dimension $d=1$, $s_{11}^{(1)}(\boldsymbol{z}_1) = h_1(w^{(1,1)}_{11}\boldsymbol{z}_1)$, $s_{11}^{(2)}(\boldsymbol{z}_1) = h_2(w^{(1,2)}_{11}s_{11}^{(1)}(\boldsymbol{z}_1) + w^{(1,2)}_{12}s_{21}^{(1)}(\boldsymbol{z}_1) + w^{(1,2)}_{13}s_{31}^{(1)}(\boldsymbol{z}_1))$. The item embeddings $\boldsymbol{S}(\boldsymbol{z}_1)$ is obtained after aggregating all $\boldsymbol{S}_d(\boldsymbol{z}_1)$ for $d = 1, 2, \dots, D$. }\label{fig:gcn}
\end{figure*}

\subsection{Generative Model}\label{sec:generative_model}
For user $u$, denote $\boldsymbol{z}_u \in \mathbb{R}^P$ as a low-dimensional latent representation of the user subgraph $\boldsymbol{A}^{(u)} \in \mathbb{R}^{K \times |\boldsymbol{V}|}$, where $P$ is the dimension of the latent feature space. For users $u_i, u_j \in \boldsymbol{U}$, the user subgraphs $\boldsymbol{A}^{(u_i)}$ and $\boldsymbol{A}^{(u_j)}$ are independent conditional on the latent embeddings $\boldsymbol{z}_{u_i}, \boldsymbol{z}_{u_j}$. Furthermore, for each user $u$, the edges $A^{(u)}_{kv}$ are conditionally independent given the latent representation $\boldsymbol{z}_u$. Therefore, the likelihood of the set of edges in $\boldsymbol{A}^{(u)}$ can be written as 
\begin{equation}\label{eq:generative}
p_{\boldsymbol{\theta}}(\boldsymbol{A}^{(u)}\mid\boldsymbol{z}_u) = \prod_{k=1}^{K} \prod_{v=1}^{|\boldsymbol{V}|} p_{\boldsymbol{\theta}}(A^{(u)}_{kv}\mid \boldsymbol{z}_u),
\end{equation}
where $\boldsymbol{\theta}$ controls the nonlinear mapping from $\boldsymbol{z}_u$ to $\boldsymbol{A}^{(u)}$ and can be learned by neural networks. 
$p_{\boldsymbol{\theta}}(\boldsymbol{A}^{(u)} \mid \boldsymbol{z}_u)$ in Equation \ref{eq:generative} is a generative model for the user subgraph $\boldsymbol{A}^{(u)}$. Specifically, we set the prior $p(\boldsymbol{z})$ as standard Gaussian, and assume that $A^{(u)}_{kv}$ are
conditionally independent Bernoulli variables with the Bernoulli parameter $f_{\boldsymbol{\theta},kv}(\boldsymbol{z}_u)$. That is, consider the generative process 

\begin{align}\label{eq:bernoulli}
&\boldsymbol{z}_u \sim N(\boldsymbol{0},\boldsymbol{I}_P), \\
&A^{(u)}_{kv} \sim \text{Bernoulli}(f_{\boldsymbol{\theta},kv}(\boldsymbol{z}_u)). 
\end{align}

There may be a set of aspects for which all users have and a set of aspects that are user-specific. To explicitly capture this, we consider a hierarchical latent space model for the nonlinear mapping  $f_{\boldsymbol{\theta},kv}$ as follows: 
\begin{align}
f_{\boldsymbol{\theta},kv}(\boldsymbol{z}_u) = \frac{e^{\gamma_{kv} + \psi_{kv}(\boldsymbol{z}_u)}}{1+ e^{\gamma_{kv} + \psi_{kv}(\boldsymbol{z}_u)}}
\label{eq:ber_param},  \\
\psi_{kv}(\boldsymbol{z}_u) = \boldsymbol{g}_k(\boldsymbol{z}_u)' \boldsymbol{s}_{v}(\boldsymbol{z}_u), \label{eq:fact}
\end{align}
where $\gamma_{kv}$ is a baseline parameter representing the shared information across all users, and $\psi_{kv}(\cdot)$ is a nonlinear function of $\boldsymbol{z}_u$, indicating the individual deviation on user $u$'s  $k$-th order proximity to item $v$. 
A positive increment on $\psi_{kv}(\boldsymbol{z}_u)$ can lead to an increment on the expectation of the connection strength for the $(k,v)$-th connection of user $u$. Intuitively, $\psi_{kv}(\boldsymbol{z}_u)$ measures user $u$'s $k$-th order preference to the $v$-th item. Therefore, we propose a latent factorization of $\psi_{kv}(\boldsymbol{z}_u)$ as an inner product between $\boldsymbol{g}_k(\boldsymbol{z}_u) \in \mathbb{R}^D$  and $\boldsymbol{s}_{v}(\boldsymbol{z}_u) \in \mathbb{R}^D$, where the former is the latent embedding of the $k$-th order proximity for user $u$, and the latter that of the $v$-th item for user $u$; see Equation \ref{eq:fact}. Intuitively, if $\boldsymbol{g}_k(\boldsymbol{z}_u)$ and $\boldsymbol{s}_v(\boldsymbol{z}_u)$ have the same sign and neither is close to zero, we have $\psi_{kv}(\boldsymbol{z}_u)>0$ leading to a positive increment on $f_{\boldsymbol{\theta},kv}(\boldsymbol{z}_u)$. 

In fact, the interactions between users and items have the  following features:  
 (1) sparsity, implying most entries in $\boldsymbol{A}^{(u)}$ are zero; (2) transitivity, implying
that item $v_1$ and item $v_2$ being connected to the $k$-th order proximity of user $u$ suggests that $v_1$ and $v_2$ are not far away
in the latent representation space; (3) distinction, meaning each item has a distinct collaborative
pattern to its neighbors; (4) similarity, implying the user-item  interactions have shared structures among users but also maintain its unique features. To fully consider the above distinct features of the user subgraphs, we leverage on measuring the item-item geometric similarity in learning $\boldsymbol{s}_{v}(\boldsymbol{z}_u)$ and exploring the high-order user-item interactions when learning $\boldsymbol{g}_k(\boldsymbol{z}_u)$.

\subsection{Learning Item and Proximity Embeddings}\label{sec:gcn} 

We aim to learn the nonlinear function  $\boldsymbol{s}_{v}(\cdot): \mathbb{R}^P \to \mathbb{R}^D$ that controls the mapping from $\boldsymbol{z}_u$ to the latent representation of item $v$ for user $u$.  Define the latent representation matrix of all items in $V$ for user $u$ as  $$\boldsymbol{S}(\boldsymbol{z}_u) = (\boldsymbol{s}_1(\boldsymbol{z}_u), \dots , \boldsymbol{s}_{|\boldsymbol{V}|}(\boldsymbol{z}_u))' \in \mathbb{R}^
{|\boldsymbol{V}|\times D}.$$ 
Each column $\boldsymbol{S}_{d}(\boldsymbol{z}_u)\in \mathbb{R}^{|\boldsymbol{V}|}$ ($d = 1, \dots, D$), the $d$-th dimensional latent representation for all items, can be viewed as an “image” with items as the irregular pixels;
and we have $D$ such “images” for each user. Therefore, the item embeddings are equivalent to the pixels representation, where the later can be learned via GCN leveraging on the geometric structure of the items.

To define the geometric structure among items, we first calculate a distance matrix of all items, and denote the distance matrix as $\boldsymbol{B} \in \mathbb{R}^{|\boldsymbol{V}|\times |\boldsymbol{V}|}$, where the entry $\boldsymbol{B}_{vv'}$ is the similarity between item $v$ and item $v'$. For each item $v$, we define its K-nearest neighbors
(K-NN($v$)) as the first K items closest to $v$ according to the distance matrix $\boldsymbol{B}$, and denote item $v$ itself as its 0-NN. Then $\boldsymbol{S}_{d}(\boldsymbol{z}_u)$ can be learned via a $M$-layer GCN defined as follows:
\begin{align*}
\boldsymbol{S}_{d}^{(1)}(\boldsymbol{z}_u) = & \,\, \boldsymbol{h}_1 (\boldsymbol{W}^{(d,1)}\boldsymbol{z}_u + \boldsymbol{b}_1) \\
\boldsymbol{S}_{d}^{(m)}(\boldsymbol{z}_u) = & \,\, \boldsymbol{h}_m( \boldsymbol{W}^{(d,m)}\boldsymbol{S}_d^{(m-1)}(\boldsymbol{z}_u) + \boldsymbol{b}_m) 
\end{align*}
for $1 \leq m \leq M$. 
$\boldsymbol{h}_m(\cdot)$ is
an activation function for the $m$-th layer, and $\boldsymbol{W}^{(d,m)}$ is the weight matrix that characterizes the convolutional operator at the $m$-th layer. For $m = 1$, $\boldsymbol{W}^{(d,1)} \in \mathbb{R}^{|\boldsymbol{V}|\times P}$. For $m \geq 2$, $\boldsymbol{W}^{(d,m)}$ is a $|\boldsymbol{V}| \times |\boldsymbol{V}|$ weight matrix, with the $v$-th row $\boldsymbol{W}_{v \cdot }^{(d,m)}$ satisfying $\boldsymbol{W}_{vv'}^{(d,m)} > 0$ if $v' = v$ or $v' \in \text{K-NN}(v),\; \text{otherwise}\; 0$. We achieve the user-specific item embedding with $\boldsymbol{S}_{d}(\boldsymbol{z}_u)= \boldsymbol{S}_{d}^{(M)}(\boldsymbol{z}_u)$. 

We also aim to learn the nonlinear function 
$\boldsymbol{g}_k(\cdot): \mathbb{R}^P \to \mathbb{R}^D$ that maps user $u$'s embedding $\boldsymbol{z}_u$ to the $k$th-order proximity embedding of user $u$. Define the latent representation matrix of all the $k$th-order proximity ($k = 1, \dots, K$) for user $u$ as
\begin{align*}
    \boldsymbol{G}(\boldsymbol{z}_u) = (\boldsymbol{g}_1(\boldsymbol{z}_u),  \dots, \boldsymbol{g}_K(\boldsymbol{z}_u))' \in \mathbb{R}^{K \times D}.
\end{align*}
Each column $\boldsymbol{G}_d(\boldsymbol{z}_u) \in \mathbb{R}^K (d = 1, \dots, D)$, the $d$-th dimensional latent embedding of the $k$-th order proximity can be learned via a $R$-layer neural network as follows: 
\begin{align*}
\boldsymbol{G}_{d}^{(1)}(\boldsymbol{z}_u)= & \,\, \tilde{\boldsymbol{h}}_1 (\tilde{\boldsymbol{W}}^{(d,1)}\boldsymbol{z}_u + \tilde{\boldsymbol{b}}_1) \\
\boldsymbol{G}_{d}^{(r)}(\boldsymbol{z}_u) = & \,\, \tilde{\boldsymbol{h}}_r( \tilde{\boldsymbol{W}}^{(d,r)}\boldsymbol{G}_d^{(r-1)}(\boldsymbol{z}_u) + \tilde{\boldsymbol{b}}_r) 
\end{align*}
for $1 \leq r \leq R$. $\tilde{\boldsymbol{h}}_r(\cdot)$ is
an activation function for the $r$-th layer, and $\tilde{\boldsymbol{W}}^{(d,r)}$ is the weight matrix that characterizes the convolutional operator at the $r$-th layer. For $r = 1$, $\tilde{\boldsymbol{W}}^{(d,1)} \in \mathbb{R}^{K \times P}$. For $r \geq 2$, $\tilde{\boldsymbol{W}}^{(d,r)}$ is a $K \times K$ weight matrix. 
Let $\boldsymbol{G}_{d}(\boldsymbol{z}_u)= \boldsymbol{G}_{d}^{(R)}(\boldsymbol{z}_u)$ be the latent embedding of all high-order user-item interactions for user $u$. We denote $\boldsymbol{b}_m,  \boldsymbol{W}^{(d,m)}$ together with $\tilde{\boldsymbol{b}}_r, \tilde{\boldsymbol{W}}^{(d,r)}$ $(m = 1, \dots, M, r = 1, \dots, R, d = 1, \dots, D)$ and $\gamma_{kv}$ as the model parameter $\boldsymbol{\theta}$ in Equation \ref{eq:generative}.

\begin{algorithm}
\caption{XploVAE Training}
\label{algorithm1}
\begin{algorithmic}

    \STATE \textbf{Input:} $\{\boldsymbol{A}^{(u)}\}_{u=1}^n, \{\boldsymbol{z}_u\}_{u=1}^n$, similarity matrix $\boldsymbol{B}$, mini-batch size $m$

    \STATE Randomly initialize $\boldsymbol{\theta}, \boldsymbol{\phi}$
    
    \WHILE{not converged}
        \STATE Sample a batch of $\{\boldsymbol{A}^{(u)}\}$ with size $m$
        
        \FOR{$u = 1, ..., m$}
            \STATE Sample $\boldsymbol{\epsilon}_u \sim N(\boldsymbol{0}, \boldsymbol{I}_P)$, and compute $\boldsymbol{z}_u = \boldsymbol{\mu}_{\boldsymbol{\phi}}(\boldsymbol{A}^{(u)}) + \boldsymbol{\epsilon}_u 
            \odot
            \boldsymbol{\Sigma}_{\boldsymbol{\phi}}(\boldsymbol{A}^{(u)})$ \\
            
            Compute $\boldsymbol{g}_k(\boldsymbol{z}_u)$ and $\boldsymbol{s}_v(\boldsymbol{z}_u)$ \\
            
            Compute $\hat{\mathcal{L}}_u$ using stratified negative sampling  \\  
            
            Compute the gradients  $\boldsymbol{\nabla}_{\boldsymbol{\theta}} \hat{\mathcal{L}}_u$ and $\boldsymbol{\nabla}_{\boldsymbol{\phi}} \hat{\mathcal{L}}_u$
            with $\boldsymbol{z}_u$
        \ENDFOR
        \STATE Average the gradients across the batch
    \ENDWHILE
    \STATE Update $\boldsymbol{\theta}, \boldsymbol{\phi}$ using gradients of $\boldsymbol{\theta}, \boldsymbol{\phi}$
    \STATE Return $\boldsymbol{\theta}, \boldsymbol{\phi}$

\end{algorithmic}
\end{algorithm}


\subsection{Variational Inference and Stratified Negative Sampling}

For the inference model that learns the inverse mapping from $\boldsymbol{A}^{(u)}$ to $\boldsymbol{z}_u$, we define 
$q_{\boldsymbol{\phi}}(\boldsymbol{z}_u \mid \boldsymbol{A}^{(u)})$ as a probabilistic encoder equipped with parameters $\boldsymbol{\phi}$. Recall the decoder is the generative model defined in Equation \ref{eq:generative}. 
The log-likelihood of $p_{\boldsymbol{\theta}}(\boldsymbol{A}^{(u)})$ is generally intractable, but can be lower bounded as 
\begin{align}
 \log p_{\boldsymbol{\theta}}(\boldsymbol{A}^{(u)}) & \geq E_{\boldsymbol{z}_u \sim q_{\boldsymbol{\phi}}}[\log p_{\boldsymbol{\theta}}(\boldsymbol{A}^{(u)} \mid \boldsymbol{z}_u)] \\ & - \text{KL}[q_{\boldsymbol{\phi}}(\boldsymbol{z}_u \mid \boldsymbol{A}^{(u)})\,||\, p(\boldsymbol{z}_u)] 
:= - \mathcal{L}_u,\label{eq:elbo}
\end{align}
based on Jensen’s inequality and a variational approximation. 
Equation \ref{eq:elbo} is referred to as the evidence lower bound (ELBO) \cite{kingma2014auto}. Therefore, our training objective is to minimize the negative of the ELBO, \textit{i.e.,} minimizing $\mathcal{L}_u$. $\mathcal{L}_u$ is comprised of two parts: the first term is the reconstruction error that measures how well
the model can reconstruct the user subgraph $\boldsymbol{A}^{(u)}$
; while the second term is a regularizer that pushes $q_{\boldsymbol{\phi}}(\boldsymbol{z}_u \mid \boldsymbol{A}^{(u)})$ to be close to $p(\boldsymbol{z}_u)$ so that we can sample it easily. We can easily add a penalizing parameter to the KL-divergence term in the ELBO to control the strength of regularization, which empirically yields improvements in model performance \cite{liang2018variational}.

Since our user subgraphs are very sparse, only the weights corresponding to items with interactions (the $1$'s) might get a significant update during the model training. The weights associated with the $0$'s would receive a marginal or no change at all. To address this problem, we adopt stratified negative sampling to reduce the number of weights updated while accounting for the "closeness" between the target user and items.
For user $u$, the log-likelihood of $\boldsymbol{A}^{(u)}$ can be written as 
\begin{align*} 
\ell_u &\equiv 
\log p_{\boldsymbol{\theta}}(\boldsymbol{A}^{(u)} \mid \boldsymbol{z}_u) \\
 &= \sum_{k=1}^{K}
\sum_{v:\boldsymbol{A}^{(u)}_{kv}=1} \{f_{\boldsymbol{\theta},kv}(\boldsymbol{z}_u)-\log (1+ e^{f_{\boldsymbol{\theta},kv}(\boldsymbol{z}_u)}) \} \\
&+ \sum_{k=1}^{K}\sum_{v:\boldsymbol{A}^{(u)}_{kv}=0}\{-\log(1+e^{f_{\boldsymbol{\theta},kv}(\boldsymbol{z}_u)})\}  = \ell_{u,1} + \ell_{u,0}.
\end{align*}

We avoid updating all weights by randomly sampling from the negative samples in the user subgraph $\boldsymbol{A}^{(u)}$. This is simple but not the best estimator of $\ell_{u,0}$, since the contribution of each $0$-connections are not homogeneous. Intuitively thinking, the latent space model assumes that nodes that are "closer" to each other are more likely to form a connection than those farther apart. To define “closeness”, we let
$
\boldsymbol{Q} \in \mathbb{R}^{|\boldsymbol{U}| \times |\boldsymbol{V}|}
$ be the shortest path length matrix where the entry $Q_{uv}$ denotes the shortest path from the user node $u$ to the item node $v$ in the incidence matrix $E \in \mathbb{R}^{\boldsymbol{U}| \times |\boldsymbol{V}|}$. Similar to stratified sampling, we divide the 0’s in $\boldsymbol{A}^{(u)}$ into $T$ strata according to $\boldsymbol{Q}$. The $t$-th stratum contains items in the set $\{v | c_{t-1} \leq Q_{uv} \leq c_t \}$, where $c_t, t = 1, \dots, T$ is a constant threshold to be specified.
We uniformly sample $n_{u,t}$ $0$'s from the $t$-th stratum which contains $N_{u,t}$ $0$'s. Our assumption that the probability of selecting items in the same stratum is the same suggests the following approximation to the log likelihood:
\begin{align*}
\hat{\ell_u} &= 
\ell_{u,1} + \hat{\ell}_{u,0} \\
&=\ell_{u,1} + 
\sum_{k=1}^K 
\sum_{t=1}^T \frac{N_{u,t}}{n_{u,t}} \sum_{v:c_{t-1} 
\leq \boldsymbol{Q}_{uv} \leq c_{t}} \{-\log (1+e^{f_{\boldsymbol{\theta},kv}(\boldsymbol{z}_u)})\},
\end{align*}
where $N_{u,t}$ is the total number of nodes with $c_{t-1} 
\leq Q_{uv} \leq c_{t}$, and $n_{u,t}$ is the number of selected samples in the $t$-th stratum. After incorporating $\hat{\ell}_u$, our VAE objective now becomes
\begin{align*}
\hat{\mathcal{L}_u} 
= - E_{\boldsymbol{z}_u \sim q_{\boldsymbol{\phi}}}[\hat{\ell_u}] + KL(q_{\boldsymbol{\phi}}(\boldsymbol{z}_u \mid \boldsymbol{A}^{(u)})\,||\,p(\boldsymbol{z}_u)). 
\end{align*}

Then, given $n$
user subgraphs, we can construct an estimator of the ELBO of the full dataset based on the mini-batches $\{\boldsymbol{A}^{(u)}\}_{u=1}^m$, a randomly drawn sample of size $m$ from the full   dataset with sample size $n$. Viewing $\frac{n}{m}\sum_{u=1}^m\hat{\mathcal{L}_u}$ as the objective, we implement a stochastic variational Bayesian
algorithm to optimize $\boldsymbol{\theta}$ and $\boldsymbol{\phi}$ respectively. Algorithm \ref{algorithm1} summarizes the XploVAE training procedure.

\section{Experiments}\label{sec:numerical}
We perform experiments on real-world datasets to evaluate
our proposed method. We aim to answer the following research questions:
\begin{itemize}
    \item \textbf{Q1:} How does XploVAE perform compared to state-of-the-art CF methods?
    \item \textbf{Q2:} How do different parameter settings (\textit{e.g.,} latent dimension size, proximity threshold, neighborhood size, \textit{etc.}) affect XploVAE?
    \item \textbf{Q3:} How does XploVAE benefit from its sub-components (\textit{e.g.,} the high-order proximity, graph convolutional layers, and stratified negative sampling)? 
\end{itemize}

\subsection{Experimental Settings}

\subsubsection{\textbf{Dataset Description} }
To evaluate the effectiveness of XploVAE, we conduct experiments on three datasets:
MovieLens-10M (ML-10M) \footnote{\url{https://grouplens.org/datasets/movielens/}}, 
Netflix \footnote{\url{http://academictorrents.com/}}, and 
Alibaba's e-commerce dataset (Alibaba) \footnote{\url{https://tianchi.aliyun.com/dataset/dataDetail?dataId=46}}, 
which are publicly accessible and vary in domain, size, and sparsity. We summarize the statistics of these datasets in Table \ref{table:1}.

\begin{itemize}
    \item \textbf{ML-10M:} These are user-movie ratings collected from a movie recommendation service. We binarize the data by keeping ratings of four or higher. To ensure the
quality of the dataset, we retain users and items with at least ten interactions.
    \item \textbf{Netflix:} This is the user-movie ratings data from the Netflix Prize competition. We binarize the data by keeping ratings of four or higher. We retain users and items with at least ten interactions to ensure data quality.
    \item \textbf{Alibaba:} This is a large user behavior dataset from Alibaba e-commerce platform. We binarize the data by keeping click and purchase behaviors. Similarly, we keep users who have clicked or purchased at least 10 items and items which have been accessed by at least 10 users.
\end{itemize}

For each dataset, we randomly select 80\% of historical user-item
interactions to constitute the training set, and treat
the remaining as the test set. From the training set, we randomly
select 10\% of interactions as the validation set to tune hyper-parameters.

\subsubsection{\textbf{Baselines} }
To demonstrate the effectiveness, we compare with the following approaches:
\begin{itemize}
    \item $\textbf{NCF}$ \cite{he2017neural}: The method  resorts to matrix factorization but replaces the inner product on the latent features of users and items with multi-layer perceptron to capture the nonlinear feature interactions.
    \item $\textbf{Mult-VAE}$ \cite{liang2018variational}: This method learns representations of observed user-item interactions using VAE and predicts users' preferences for items using a generative model with multinomial likelihood. 
    \item $\textbf{INH-MF}$ \cite{lian2018inhmf}: It utilizes a matrix-factorization-based information network hashing algorithm to learn binary codes which can preserve the high-order proximity.
    \item $\textbf{NGCF}$ \cite{xiang2019ngcf}:
    This is a state-of-the-art GCN-based recommender model that integrates the second-order proximity between users and items into a bipartite graph structure by propagating embeddings with the message-passing mechanism on the user-item interaction graph.
    \item \textbf{T(J \& P)} \cite{Nakatsuji2010music}: It employs a graph-based approach to accurately identify items for the target user by analyzing the interests of users who share same items with the target user and identify items with higher novelty for the user.
\end{itemize}

\begin{table}[ht!]
  \begin{center}
  \resizebox{.8\linewidth}{!}{%
    \begin{tabular}{ l c c c  }
      \toprule    
       &  ML-10M & Netflix & Alibaba  \\ 
    \midrule
    Users &  69,838 & 65,533 & 49,235 \\
Items &  8,940 &  17,759 & 234,647 \\
Interactions &  10.0M &  25.0M & 56.0M\\
Density &  0.23\% & 2.15\% & 0.02\% \\
      \bottomrule 
    \end{tabular}
  }
  \caption{Statistics of the datasets. Interactions is the number of non-zero entries in the user-item matrix, and density is the proportion of non-zero entries.}\label{table:1}
  \end{center}
\end{table}
 
\subsubsection{\textbf{Evaluation Methods} }
The effectiveness of recommendation models is conventionally assessed with relevance metrics such as NDCG or Recall at K. However, relevance alone is not a clear indicator of the quality of the recommended items. A good recommendation method should also consider exploitation of the user profile and exploration of novel products. We propose to evaluate XploVAE on two aspects: \textit{relevance} and \textit{exploration}.
The \textit{relevance} aspect gives priority to items that have
high predicted rating. The \textit{exploration} aspect gives priority to novel items that are outside of the user’s past preferences. Given that the user profile can be incomplete, it is smart to retrieve information on unknown user preferences.

After model training and validation, we evaluate our method and the competitors on the test set using the following evaluation metrics. We report the
average metrics for all users in the test set.  
\begin{itemize}
    \item \textbf{NDCG@K}: The normalized discounted cumulative gain estimates the quality of recommendation accuracy for all items ranked within the first $K$. For each user $u$, the truncated discounted cumulative gain (DCG@K) is
    $$
    \text{DCG@K}(u, \boldsymbol{w}) = \sum_{k=1}^K \frac{2^{\mathbb{I}[w(k)\in \boldsymbol{I}_u]} - 1}{\log(k+1)}, 
    $$
    where $\boldsymbol{w}$ is the ranked list of recommended items, $w(k)$ is the item at rank $k$, $\boldsymbol{I}_u$ is the set of all held-out items that the user $u$ clicked on, and $\mathbb{I}[\cdot]$ is an indicator function. NDCG@K is the DCG@K normalized to $[0, 1]$ after dividing by the best possible DCG@K.
    \item  \textbf{Recall@K}: Recall at K is the proportion of relevant items found in the top-K recommendations. Formally, Recall@K is defined as
    $$
    \text{Recall@K}(u, \boldsymbol{w}) = \frac{\sum_{k=1}^{K} \mathbb{I}[w(k) \in \boldsymbol{I}_u]}{\min(K, |\boldsymbol{I}_u|)}.
    $$
    \item \textbf{PILD} \cite{barraza2017xplodiv}: The diversity of the recommendation list is measured as the pairwise inter-list dissimilarity (PILD) of the items. Mathematically, PILD is 
    $$
    \text{PILD}(u, \boldsymbol{w}) = \frac{2}{|\boldsymbol{w}|(|\boldsymbol{w}|-1)}\sum_{i=1}^{|\boldsymbol{w}|} \sum_{j=i+1, j\neq i}^{|\boldsymbol{w}|}d(w(i), w(j)),
    $$
    where the distance $d(w(i), w(j)) = 1 - \text{sim}(w(i), w(j))$  and we use Jaccard similarity coefficient for the similarity between items. 
\end{itemize}

\subsubsection{\textbf{Parameter Settings}} We implement our proposed method in
Pytorch. We select model hyper-parameters and architectures by evaluating Recall@20 on the validation users. For XploVAE, $c_k$ in Definition \ref{def: subgraph} is fixed to 0.9 in user subgraph construction. We set the dimension of user embeddings $P$ to 200 and the dimension of item and proximity embeddings $D$ to 3. The neighborhood size $K$ for learning the geometric structure of items is 300. We also introduce a penalty parameter $\gamma$ with a value of $0.2$ to control the strength of regularization on the KL divergence term of the ELBO \cite{liang2018variational}. The parameters of the baseline methods are  suggested by the original paper. We optimize
all models with the Adam optimizer, where the
batch size is fixed at 512. We apply a
grid search for hyper-parameters: the learning rate is tuned to
0.001 and the dropout ratio is 0.1. Xavier initialization \cite{xavier2010init} is used to initialize the model parameters. We do not apply weight decay for any models. Early stopping strategy is adopted and 1000 epochs are typically sufficient for
XploVAE to converge. We
keep the model with the best validation NDCG@20 and report
test set metrics with it.

\begin{figure}[ht!]
\centering
\includegraphics[width=.47\textwidth]{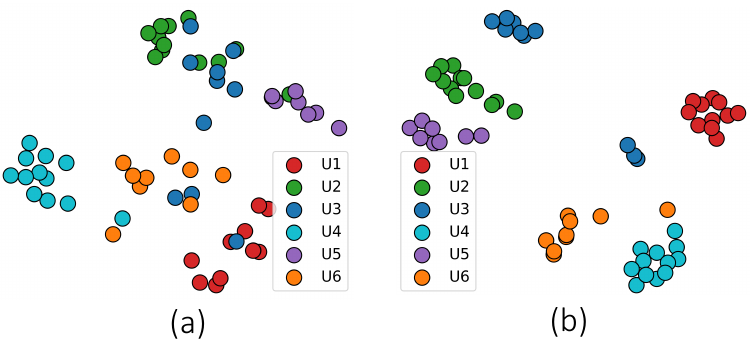}
\caption{Visualization of the learned t-SNE 
representations derived from (a) XploVAE-1st that only includes the first-order proximity and (b) XploVAE that includes both the first-and  second-order proximity. The points with the same color denote relevant items accessed by the same user.}\label{fig:TSNE}
\end{figure}

\begin{figure*}[ht!]
\centering
\includegraphics[width=0.9\textwidth]{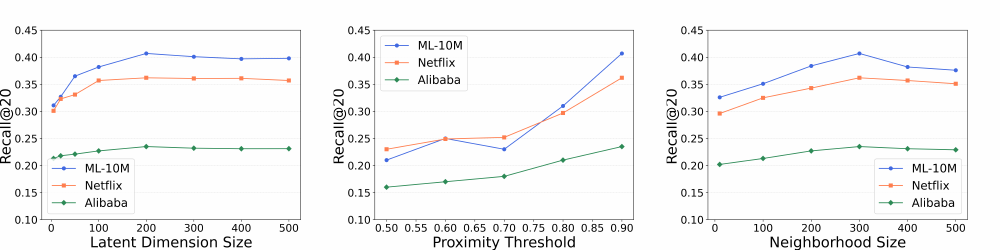}
\caption{Performance of XploVAE w.r.t different latent dimension size, proximity threshold and neighborhood size across the three benchmark datasets.}\label{fig:parameter_sensitivity}
\end{figure*}

\subsection{Results and Analysis}

\subsubsection{\textbf{Comparative Study (Q1)}}
To demonstrate the effectiveness, we compare our proposed XploVAE to the baseline methods. The effectiveness of recommender systems is assessed with relevance metrics such as Recall@K and NDCG@K. By default, we set K = 20. Table \ref{table:2} and \ref{table:3} report the performance comparison results. We have the following findings:
\begin{itemize}
    \item Among our competitors, we observe that methods preserving the high-order connectivity, including INH-MF and NGCF, recommend items of higher relevance in comparison to NCF and Mult-VAE, which only exploit the first-order proximity. 
    \item T(J \& P) generally underperforms the other CF methods, indicating that methods that only consider exploration fail to fully explore the nonlinear feature interactions between users and items.
    \item  NGCF outperforms NCF, Mult-VAE, INH-MF and T(J \& P) in all cases. Such improvements might be attributed to the explicit modeling of crucial collaborative signals in the embedding function.
    \item XploVAE consistently achieves the best performance on all benchmark datasets of varying type, size and sparsity. This verifies the importance of modeling the high-order connectivity, learning personalized item embeddings from user subgraphs and capturing collaborative signals in the item embedding function. The comparatively low NDCG@20 and Recall@20 scores on the Alibaba dataset indicate that XploVAE yields more accurate link predictions for small and medium-sized datasets, but performs less desirably on large datasets.
\end{itemize}

We also attempt to understand how the inclusion of user-specific item embeddings facilitates the representation learning in the
embedding space. Towards this end, we randomly selected six users
from the Alibaba dataset, as well as their relevant items. We observe
how their representations are influenced w.r.t. having shared or individual item embeddings. Figure \ref{fig:TSNE} (a) and (b) depict the visualizations of the representations derived from XploVAE-1st that only includes the first-order proximity and XploVAE that includes both the first- and second-order proximity, respectively. A key observation is that items belonging to the same user are embeded into the near part of the space. In particular, the representations of XploVAE exhibit discernible clustering, meaning that points with the same colors (\textit{i.e.,} items purchased by the same users) tend to form the clusters. It qualitatively verifies that learning user-specific item embeddings is capable of making more personalized recommendations.

A good recommendation model should also consider exploration of novel products in addition to recommending relevant items. We use PILD as the exploration metric to evaluate our method's ability to recommend novel items. The results are shown in Table \ref{table:4}. We observe that methods that consider exploration (\textit{e.g.,} INH-MF, NGCF, T(J \& P) and XploVAE) generally achieve better performance w.r.t PILD, suggesting that items recommended by these methods are more diverse. For all datasets, XploVAE consistently scores the highest in terms of PILD among all competing approaches. Compared to other methods, XploVAE recommends more diverse items to users while ensuring the items recommended are highly relevant. 

\begin{figure*}[ht!]
\centering
\includegraphics[width=0.9\textwidth]{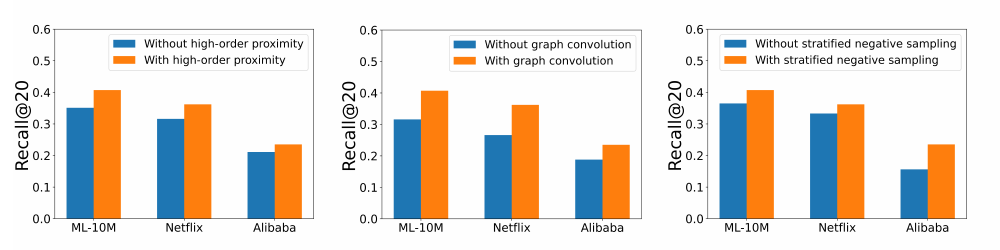}
\caption{Results of XploVAE and its variants that do not include the designed components across the three benchmark datasets.}\label{fig:ablation study}
\end{figure*}

\begin{table}[ht!]
  \begin{center}
  \resizebox{.75\linewidth}{!}{%
    \begin{tabular}{ l   c  c  c  }
    \toprule
       &  ML-10M & Netflix & Alibaba  \\ 
    \midrule
    NCF & 0.360 & 0.316  & 0.188\\
    Mult-VAE & 0.387 & 0.344 & 0.211 \\
INH-MF & 0.370 & 0.337 & 0.206 \\
NGCF & 0.395 & 0.351 & 0.220\\
T(J \& P) & 0.323 & 0.286 & 0.157\\
\midrule
\textbf{XploVAE} & \textbf{0.407} & \textbf{0.362} & \textbf{0.235}\\
\bottomrule 
    \end{tabular}
  }
  \caption{Performance comparisons for benchmark datasets based on the relevance metric Recall@20.}\label{table:2}
  \end{center}
\end{table}

\begin{table}[ht!]
  \begin{center}
  \resizebox{.75\linewidth}{!}{%
    \begin{tabular}{ l   c  c  c  }
    \toprule
       &  ML-10M & Netflix & Alibaba  \\ 
    \midrule
    NCF & 0.352 & 0.297 & 0.158 \\
    Mult-VAE & 0.379 & 0.323 & 0.184 \\
INH-MF & 0.357 & 0.330 & 0.165 \\
NGCF & 0.384 & 0.339 & 0.201 \\
T(J \& P) & 0.315 & 0.277 & 0.126 \\
\midrule
\textbf{XploVAE} & \textbf{0.391} & \textbf{0.343} & \textbf{0.217} \\
\bottomrule 
    \end{tabular}
  }
  \caption{Performance comparisons for benchmark datasets based on NDCG@20.}\label{table:3}
  \end{center}
\end{table}

\subsubsection{\textbf{Parameter Sensitivity (Q2)}}

\begin{table}[ht!]
  \begin{center}
  \resizebox{.75\linewidth}{!}{%
    \begin{tabular}{ l   c  c  c  }
    \toprule
       &  ML-10M & Netflix & Alibaba  \\ 
    \midrule
    NCF & 0.209 & 0.318 & 0. 562 \\
    Mult-VAE & 0.211 & 0.329 & 0.539 \\
INH-MF & 0.227 & 0.334 & 0.543 \\
NGCF & 0.265 & 0.351 & 0.621 \\
T(J \& P) & 0.244 & 0.346 & 0.627 \\
\midrule
\textbf{XploVAE} & \textbf{0.296} & \textbf{0.375} & \textbf{0.675} \\
\bottomrule 
    \end{tabular}
  }
  \caption{Performance comparisons for benchmark datasets based on PILD.}\label{table:4}
  \end{center}
\end{table}

In this part, we study the effects of the following hyper-parameters on the relevance metric: 
\begin{itemize}
    \item $P$: The dimension of the latent representation $\boldsymbol{z}_u$ as defined in Section \ref{sec:generative_model}.
    \item $c_k$: The proximity threshold in user subgraph construction as defined in Definition \ref{def: subgraph}.
    \item $K$: The neighborhood size for learning the item embeddings in Section \ref{sec:gcn}.
\end{itemize}
Specifically, we consider the variants of XploVAE equipped with different hyper-parameter settings. The results are shown in Figure \ref{fig:parameter_sensitivity} and we have the following findings:
\begin{itemize}
    \item Increasing the latent dimension size enhances the recommendation cases at first. The performance of XploVAE starts to decline after reaching the plateau at a dimension size of 200. 
    \item Similarly, when increasing the neighborhood size, we observe an improvement in the performance of XploVAE. A slight drop in the relevance score is also observed after peaking at a neighborhood size of 300.  
    \item Increasing the proximity threshold substantially increases the relevance of recommended items. 
    \item Relatively large fluctuations in Recall@20 due to different hyper-parameter settings are observed for the ML-10M and Netflix datasets, while the discrepancies are small for the Alibaba dataset. This suggests that the performance of XploVAE on small datasets is more sensitive to changes in hyper-parameters. On the contrary, link prediction tasks for large datasets such as Alibaba are inherently hard and changing the model hyper-parameters does not noticeably affect the model performance. 
\end{itemize}

\subsubsection{\textbf{Ablation Study (Q3)}}
To investigate whether XploVAE can benefit from the adaptation of the high-order proximity, graph convolution and stratified negative sampling, we perform ablation studies to verify the effects of these components on the model performance. We show the results of XploVAE and its variants in Figure \ref{fig:ablation study} and have the following findings:
\begin{itemize}
    \item Explicitly modeling the high-order connectivity is consistently superior to considering the first-order proximity only across all benchmark datasets. We attribute the improvements to the potential interactions captured by the transitive associations between users and items, which allows us to explore user interests while not over-exploiting known user profiles.
    \item Exploiting the graph convolution to leverage item-item similarity substantially enhances the recommendation relevance. Clearly, XploVAE equipped with graph convolutional layers achieves
consistent improvement over vanilla XploVAE, which does not inject collaborative signals into item embeddings, across all datasets. We attribute the improvement to the effective modeling of the collaborative item-item similarity. 
    \item Including stratified negative sampling in XploVAE yields only marginally superior performance on the ML-10M and Netflix datasets. However, we observe a noticeable improvement in the relevance metric on the Alibaba dataset with stratified negative sampling. We attribute such a discrepancy to the larger sparsity of the Alibaba dataset.
\end{itemize}

\subsubsection{\textbf{Efficiency Study}}
\begin{figure}[ht!]
\centering
\includegraphics[width=0.47\textwidth]{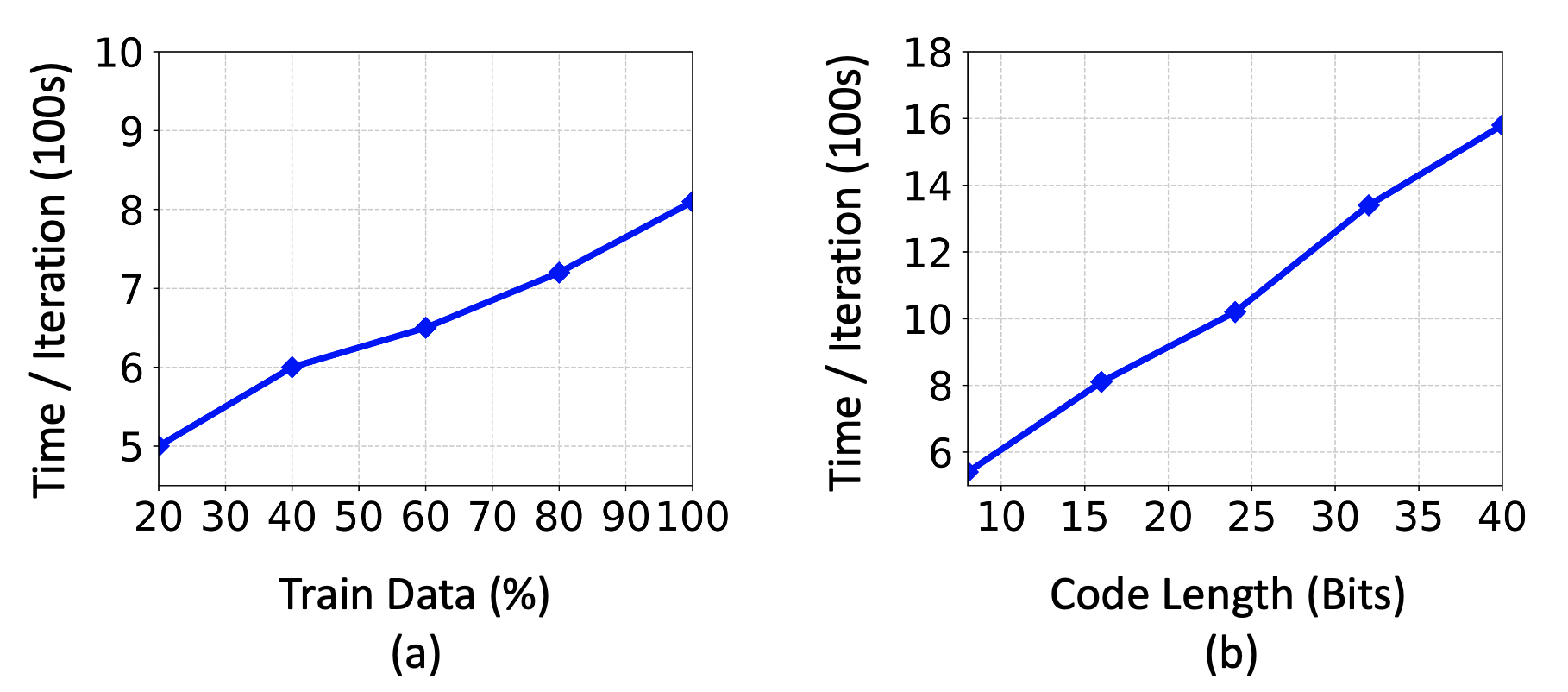}
\caption{Efficiency w.r.t. data size and code length on Netflix.}\label{fig:efficiency}
\end{figure}

We finally investigate the efficiency of XploVAE with the increase of code length and training data size, and depict the results in Figure \ref{fig:efficiency}. The experiments are executed on the Netflix dataset. Firstly, we fix the code length as 8 bits, and gradually increase the size of training data. Secondly, we fix the training ratio (50$\%$ for training), and then tune the value of code length. In both cases, the time required per iteration will be linearly increased, indicating the potential to be applied in large-scale datasets. XploVAE is comparably efficient to other recommender systems methods as the competitor approach INH-MF also scales linearly with training data size and almost quadratically with code length \cite{lian2018inhmf}. As a result, XploVAE is efficient and suitable for recommendation problems.

\section{Conclusion}\label{sec:summary}

In this paper, we propose a variant of VAE denoted as XploVAE that utilizes exploitation of user profiles and exploration of novel items. We construct user subgraphs from the user-item interaction history with the first-order and the high-order proximity for personalized recommendations. The hierarchical latent space model is integrated into the VAE framework, allowing us to learn the latent representations of individual user subgraphs along with the population distribution of all subgraphs. Experimental results show the effectiveness of our approach, which outperforms the existing collaborative filtering methods that adopt a VAE framework or preserve the high-order connectivity.

This work represents an initial attempt to learn personalized embeddings via the construction of individual user subgraphs. It also opens up new research opportunities to integrating the high-order proximity into VAE-based models for exploration purposes. In the future, we plan to investigate efficient and inductive \cite{hamilton2017inductive} ways of learning item embeddings via GCN so that XploVAE can scale up to very large networks.

\bibliographystyle{ACM-Reference-Format}
\bibliography{bibfile}


\end{document}